\documentclass[letterpaper, 10 pt, conference]{ieeeconf}  

\IEEEoverridecommandlockouts                              

\usepackage{cite}
\usepackage{amsmath,amssymb,amsfonts}
\usepackage{algorithmic}
\usepackage{graphicx}
\usepackage{subfig}
\usepackage{textcomp}
\usepackage{float}
\usepackage{tabularx}
\usepackage{subfig}
\usepackage{multirow}
\usepackage{graphicx}
\usepackage{hhline}
\usepackage{float}
\usepackage{adjustbox}
\usepackage{float}
\usepackage{xcolor}
\usepackage{array, makecell, booktabs}
\newcommand{\RN}[1]{%
  \textup{\uppercase\expandafter{\romannumeral#1}}%
}   

\title{\LARGE \bf
Self-Guided Instance-Aware Network for Depth Completion and Enhancement
}

\author{Zhongzhen Luo$^{1}$, Fengjia Zhang$^{2}$, Guoyi Fu$^{1}$, Jiajie Xu$^{2}$ 
\thanks{$^{1}$Epson Research, Markham, Ontario, Canada.
        {\tt\small \{zhongzhen.luo, gary.fu\}@ea.epson.com}}%
\thanks{$^{2}$University of Toronto, Toronto, Ontario, Canada.
        {\tt\small \{fuuka.zhang, jeremy.xu\}@mail.utoronto.com}}%
}

\begin{document}

\maketitle
\thispagestyle{empty}
\pagestyle{empty}

\begin{abstract}

Depth completion aims at inferring a dense depth image from sparse depth measurement since glossy, transparent or distant surface cannot be scanned properly by the sensor. Most of existing methods directly interpolate the missing depth measurements based on pixel-wise image content and the corresponding neighboring depth values. Consequently, this leads to blurred boundaries or inaccurate structure of object. To address these problems, we propose a novel self-guided instance-aware network (SG-IANet) that: (1) utilize self-guided mechanism to extract instance-level features that is needed for depth restoration, (2) exploit the geometric and context information into network learning to conform to the underlying constraints for edge clarity and structure consistency, (3) regularize the depth estimation and mitigate the impact of noise by instance-aware learning, and (4) train with synthetic data only by domain randomization to bridge the reality gap. Extensive experiments on synthetic and real world dataset demonstrate that our proposed method outperforms previous works. Further ablation studies give more insights into the proposed method and demonstrate the generalization capability of our model. 

\end{abstract}

\section{INTRODUCTION}

Depth sensor plays a crucial role in many robotics applications that require an interpretation of the scene. For example, \cite{kp-gc} use semantic keypoints as object representation to plan robot trajectories by using one of high-end commodity-level depth sensors. In spite of the recent advances in depth sensing technology, object surfaces such as shiny, glossy, transparent, and distant pose challenges for depth measurements as shown in Fig. 1. One promising attempt is to alternatively capture the images sequentially in different orientations, combining multiple views, and manually annotate ground truth data. However, such solution requires prohibitively expensive cost and suffers from dynamic objects and high latency, which makes it a less practical solution for real-time robotics applications. Towards this end, the goal of our work is to develop an affordable and efficient solution for depth completion that only train on synthetic data, and test with a single view of a commodity-level RGB-D camera.

\begin{figure}[ht]
\begin{center}
\includegraphics[width=1\linewidth]{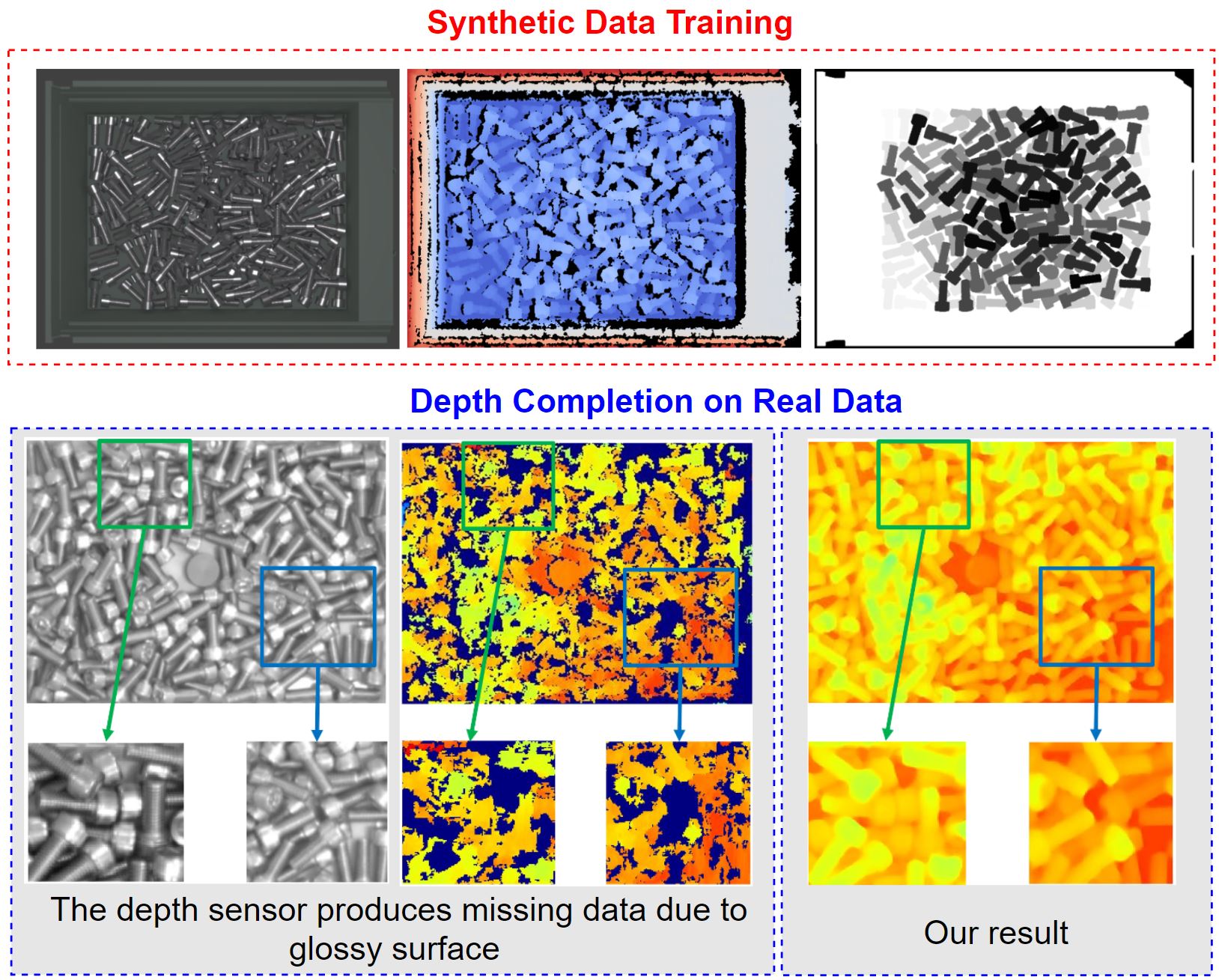}
\end{center}
   \caption{Sample illustration of our approach. Our approach is trained on synthetic data only. At test time, the same model is used in real world with no additional training. }
\end{figure}

Traditional methods to estimate the missing depth values can be achieved by explicitly using the hand-tuned methods \cite{hand1,hand2}, or finding the local affinity or discontinuity based on interpolation and diffusion schemes \cite{hand3}. With the recent explosive growth of deep learning techniques, several methods have been introduced into this task and reveal promising improvements \cite{cnn1,cnn2,cnn5}. However, they basically share the same scheme by directly using the element-wise addition operation for feature fusion. Applying such a simple way leads to blurred boundaries and inaccurate shape reconstruction. Recent approaches \cite{cnn3} aim to mitigate these failures with local representation prediction from color images, and solve for depth completion via a traditional optimization problem. But we observe that such multi-stage algorithms often fail easily in cluttered scenarios where edges and instances are ambiguous.

In contrast, as convolutional neural network (CNN) have been successful in learning effective representations in object detection \cite{det1,det2,det3}, and pose estimation \cite{pose1,pose2}. These works demonstrate that CNN are very powerful for understanding semantic context, which inspire us the study of instance level features. One of our observations from this perspective is that the geometric property that position relationship between points of a rigid instance in 3D space is fixed. Inspired by the benefits of keypoint voting in \cite{pose3}, we argue that to address occlusion problems requires pixel-wise estimation of directions from each pixel towards the instance keypoints (e.g., center, corner, distinguished feature). Secondly, inspired by the success of guided image filtering \cite{gate1}, we introduce an unsupervised self-guided module, to learn the dynamic feature for each channel and each spatial location, hence the proposed methods will pay more attention on relevant information. Furthermore, we leverage the negative log-likelihood of the observed depth as uncertainty into the cost term, which not only force the network to find semantic clues from color information, but also enable the network to attenuate the effort on the observed region. In addition, we handle the issues of vague structures and multiple objects with context information concepts and global geometric constraint. Instead of predicting depth values only, we equip our network to concentrate on instance-aware consistency via joint tasks learning. By jointly optimizing with instance-aware tasks, we find that it can boost the performance of each other.

To summarize, the main contributions of our work are the followings: (1) We develop a novel end-to-end network for depth completion and enhancement by jointly considering the instance-level features, and utilizing geometric constraint and context information to deals with occlusion and vague structures. (2) We also propose to use unsupervised self-guided mechanism together with negative log-likelihood function to learn a dynamic feature selection for each channel at each spatial location of layers to enhance the depth completion performance. (3) We present domain randomization to bridge the gap between synthetic and real so that we can train our model by using synthetic data only. (4) We demonstrate improvements of our approach compared to the state-of-the-art (SOTA) approaches on both synthetic and real KITTI datasets. Because our proposed method train on synthetic data only, we also evaluate on our real world in-house robotics dataset to demonstrate the practical ability for robustness and generalization.

\section{Related Work}
Depth completion has been intensively studied since the emergence of active depth sensors. We briefly review these techniques and other literatures relevant to our network design.

\begin{figure*}[ht]
\begin{center}
\includegraphics[width=0.99\linewidth]{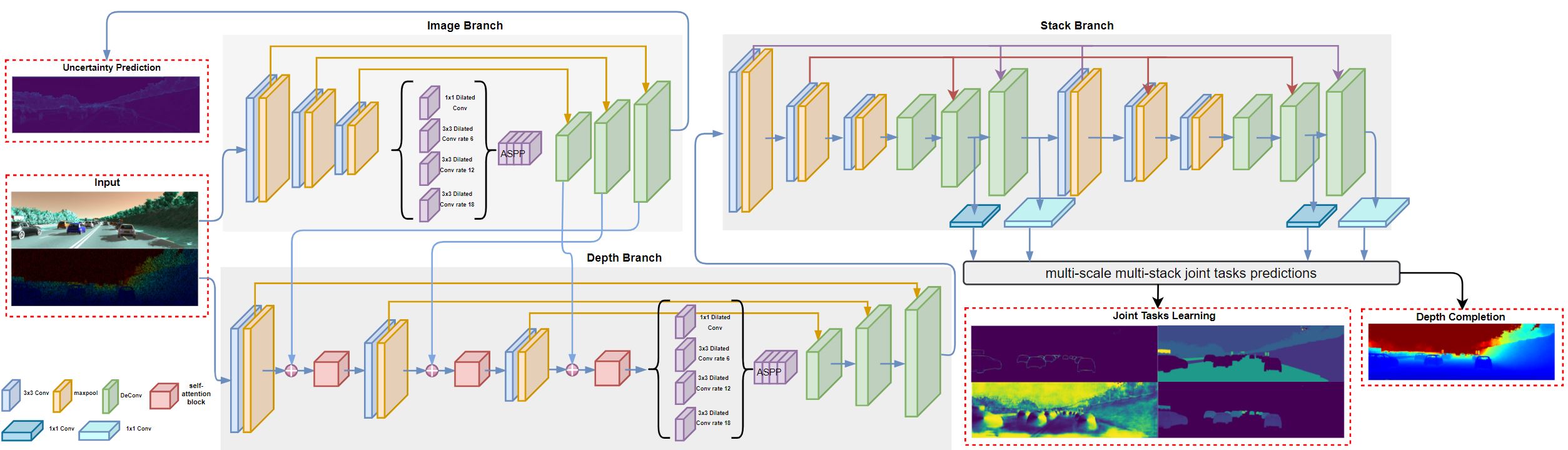}
\end{center}
   \caption{Overall pipeline of our proposed method: The image branch takes image as input, and the depth branch takes depth as input. The stack branch takes the fusion features from depth branch and performs coarse-to-fine multi-scale joint tasks learning.}
\end{figure*}

\subsection{Image-only method}
The topics gained popularity recently by using neural networks to learn depth information from pictorial cues. Eigen et.al was the first to use multi-scale neural network to achieve coarse-to-fine prediction \cite{imgcnn1}. Since then, various learning-based approaches were developed based on monocular color images \cite{imgcnn2,imgcnn3,mono3}. Yang et al. \cite{imgdep2} introduced surface normal representation for geometric constrain. Built upon this, they then further introduced edge consistency in parallel with surface normal consistency for fine detailed structures recovery \cite{imgdep3}. Recent work in this field focus on improving the prediction accuracy and reducing pixel relative error through various methods, such as depth discretization (DORN) \cite{dorn}, 3D geometric constraints (VNL) \cite{vnl}, and local planar guidance (BTS) \cite{bts}. These methods share some ideas with our work. However, the motivation of these works is usually to compensate the constraints on camera size, cost and image quality. And therefore, they are not suitable for predicting accurate pixel level values on semi-dense depth measurement, specifically on the pixels that are missing from raw depth measurement.

\subsection{Depth with image method}
Several approaches attempted to use guidance from additional image input. Eigen et al. \cite{cnn9} proposed to use CNN for multitasks: depth, surface normal and semantic segmentation. Laina et al. \cite{cnn10} used fully CNNs, encompassing residual learning architecture to model the mapping between monocular images and depth maps. Another sub-topic in this area focused on reconstructing dense depth map by augmenting sparse depth map with single color image. Ma et al. \cite{cnn7} first introduced this method and achieved good performance on KITTI dataset. Ma et al. \cite{cnn8} further improved this method to train without semi-dense annotations, while still outperformed previous supervised training solutions. Recent depth completion research focus on fixing the missing area from relatively dense RGB-D data. Zhang et al. \cite{cnn3} extracted surface normal and occlusion boundary as the feature representation from RGB data. This effectively conquer the issue that the traditional deep regression method simply learns to copy or interpolate values from the context. Build upon this work, Huang et al. \cite{cnn4} further emphasize the depth reconstruction on object boundaries, by adding boundary consistency into the pipeline. However, their work still lack of global features, and fail easily in cluttered scenarios.

\section{Proposed Method}
Given the raw depth input $D$, the learning-based depth completion can be summarized as the process of seeking to find a predictor $f_{\theta}$ such that the predicted depth map $\hat{D}$ is as close as possible to the ground truth depth map $D_{gt}$. Formally, the depth completion process is formulated to minimize the objective function $\mathcal{L}$ of the form as: $\mathcal{L} = \sum_{k=1}^{n} C(f_{\theta}(I_k,D_k),D_{gt}) $, where $I_k$ and $D_k$ are the pixel-wise image and depth value respetively. $C(,)$ is a certain measure of distance between the ground truth depth map and the predicted depth map, and $n$ is the sets of image pixels of the same scene.

\subsection{Network Overview}
To tackle the task as fomulated above, we propose an end-to-end framework that takes an RGB image and a depth image as inputs and produce a complete depth image. The overall architecture is illustrated at Fig. 2. Inspired by recent methods \cite{codeslam}, the whole network mainly consists of three branches, the image branch, depth branch and stack branch, as described in the following subsections. 

\subsubsection{\textbf{image branch}}
The goal of the image branch is to compute an increasingly coarse but high-dimensional feature representation of the image by a series of ResNet blocks. This is followed by an Atrous Spatial Pyramid Pooling (ASPP) module \cite{guide2} with dilated convolution for effective incorporation of hierarchical context information. Then the up-sampling part is equipped with skip-connection layers and shared pooling masks for corresponding max pooling and unpooling layers. To learn a function over the prediction error, we employ an uncertainty map, which consists of training the network to infer mean and variance of the distribution $\textbf{P}(D_{gt}|D,I)$ of parameters $\Theta$. Hence the network is trained by log-likelihood maximization: 
\begin{equation}
    \text{log}(\textbf{P}(D_{gt}|W)) =  \frac{1}{n}\sum_{k=1}^{n} \text{log}(\textbf{P}(D_{gt}|\Theta(I, D,W)))
\end{equation}

\noindent where W is the training weights. As presented in \cite{codeslam}, the predictive distribution can be modelled as Laplace distribution, because it allows the model to attenuate the cost of non-missing regions and to focus on reconstructing on missing region via image-to-depth fusion. Therefore, we derive the cost term $\mathcal{L}_{u}$ by minimizing the following:

\begin{equation}
    \mathcal{L}_{u} =  \frac{|\mu(D)-D_{gt}|}{\sigma(D)} + \text{log}(\sigma(D))
\end{equation}

\noindent where $\mu(D)$ and $\sigma(D)$ are the outputs of the model encoding mean and variance of the distribution, while being subject to the additional logarithmic term to discourage infinite predictions. Moreover, the learned uncertainty map can also serve to gauge the reliability of depth completion at run time.

\subsubsection{\textbf{depth branch}} The depth branch is to fuse the features from depth and image features together. Commonly used copy or interpolation operation can easily fall into local minima instead of predicting precise depth values. To avoid this problem, we proposed to use self-guided module, as shown in red box of Fig. 2, in depth branch, such that the network is self-guided to forward important features by paying more attention not only to global features, but also to instance level features. Therefore, the fusion module allows the depth branch to learn dynamic feature selection for each channel and each spatial location. The self-guided module is formulated as:

\begin{equation}
    F_{out} = \sigma(\sum_{k=1}^{n} W_g \otimes (D \uplus I)) \odot \delta(\sum_{k=1}^{n} W_f \otimes (D \uplus I)) \oplus D
\end{equation}

\noindent where $D$ and $I$ are features from depth and image branch. $\uplus$ represents concatenation operation of layers, $\otimes$ represents convolution operation, $\oplus$ represents element-wise summation and $\odot$ denotes the element-wise multiplication. While $\sigma$ is sigmoid function, $\delta$ can be any other activation functions, such as ReLU, Linear or Tanh. $W_g$ and $W_f$ are two different convolutional filters for each activation function.

\begin{figure}[ht]
\begin{center}
\includegraphics[width=0.9\linewidth,height=0.15\textheight]{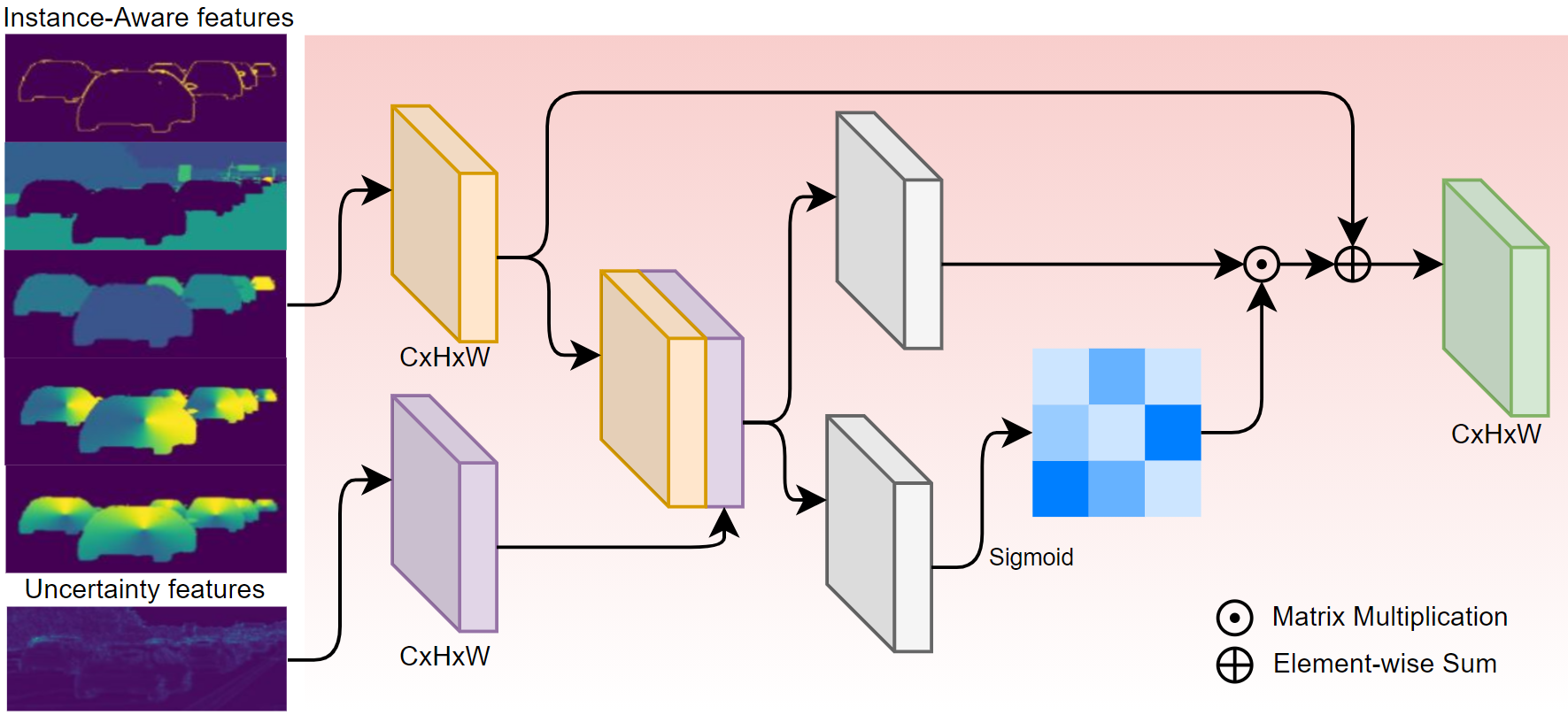}
\end{center}
   \caption{Sample of instance-aware learning with self-guided module}
\end{figure}

\subsubsection{\textbf{stack branch}} The stack branch is designed to consist of repeated encoder-decoder architecture with intermediate output. This allows the network for re-evaluation and re-assessment of initial estimates and helps to maintain precise local information while considering and then reconsidering the overall coherence of the features. With such stacked hourglass architecture, our network performance has been improved compared to baseline as shown in the ablation study in Section \RN{4}-D. 

\subsection{Instance-Aware Learning} 
Leveraging the complementarity properties of geometry and context information for instance-aware features, we propose to jointly solve these tasks so that one boosts the performance of another as shown in Fig. 3. Instead of a distillation multi-module proposed by \cite{xu}, we propose to split the tasks at the last layers so that all the tasks share knowledge through the network and guided by the self-guided module.

Firstly, we propose to learn the instance center consistency by localizing the object center in the image and estimating object distance from the camera. We predict pixel-wise unit vectors that represent the direction from each pixel to center of the object. Predicting pixel-wise directions alleviates the influence of cluttered background since the invisible part of the object can be correctly located from other visible parts in terms of the directions. In addition, we use object distance label to strengthen the network ability to handle object with similar shape but different depth values. Therefore, our first task is to preserve center consistency $\textbf{T}$ such that: $\textbf{T} = (v_x = \frac{p_x-c_x}{|\textbf{p}-\textbf{c}|},v_y = \frac{p_y-c_y}{|\textbf{p}-\textbf{c}|}, T_z)$, where $p_x$ and $p_y$ are 2D pixel position in image. $c_x$ and $c_y$ are instance center position, and $T_z$ is the semantic label for instance distance. To handle sensor noise or inaccurate depth measurement, we adopt smooth L1 loss terms $\mathcal{L}_1$ for the task of center consistency. 

Secondly, our model predict semantic information to handle the foreground and background differentiation. Given the pixel-wise extracted feature, semantic segmentation task $\mathcal{L}_{ss}$ is supervised by the binary cross entropy loss. Thirdly, we also introduce object boundary consistency to enforce model to preserve the clear boundaries in the output depth, making the depth image to be more structured and conformed to realistic situation. The boundary consistency task $\mathcal{L}_{bc}$ is also supervised by binary cross entropy loss. Because depth and surface normal are two strongly correlated factors, the locally linear orthogonality between them can be utilized to regularize the depth completion. Motivated by \cite{normal1}, we also estimate the surface normal and incorporate a depth-normal consistency term for the normal guided depth completion into a loss function. The depth-normal consistency loss is defined as: $\mathcal{L}_{sn} = \sum_{p,q\subseteq n} ||<V(p,q),N(p)>|| ^2$, where $<,>$ denotes an inner product. The $\mathcal{L}_{sn}$ measures the consistency such that the inner product of the vector from point $p$ to its neighbor $q$ should be zero with surface normal vector $N(p)$. 

Therefore, in order to guide the update of the network weights to measure how far the estimated depth map $\hat{D}$ is from the ground truth depth map $D_{gt}$, we design the cost terms in our pipeline as: $\mathcal{L}_{total} = \lambda_1 \mathcal{L}_{u} +  \lambda_2 \mathcal{L}_1(\textbf{T}_{gt},\hat{\textbf{T}}) +\lambda_3 \mathcal{L}_1(D_{gt},\hat{D}) + \lambda_4 \mathcal{L}_{ss} + \lambda_5 \mathcal{L}_{bc} + \lambda_6 \mathcal{L}_{sn} $. The first two terms measure the errors between the ground truth and the estimated depth while the rest instance-aware task terms are used to incorporate various geometry and context constraints.

\begin{figure}[ht]
\begin{center}
\includegraphics[width=1\linewidth,height=0.22\textheight]{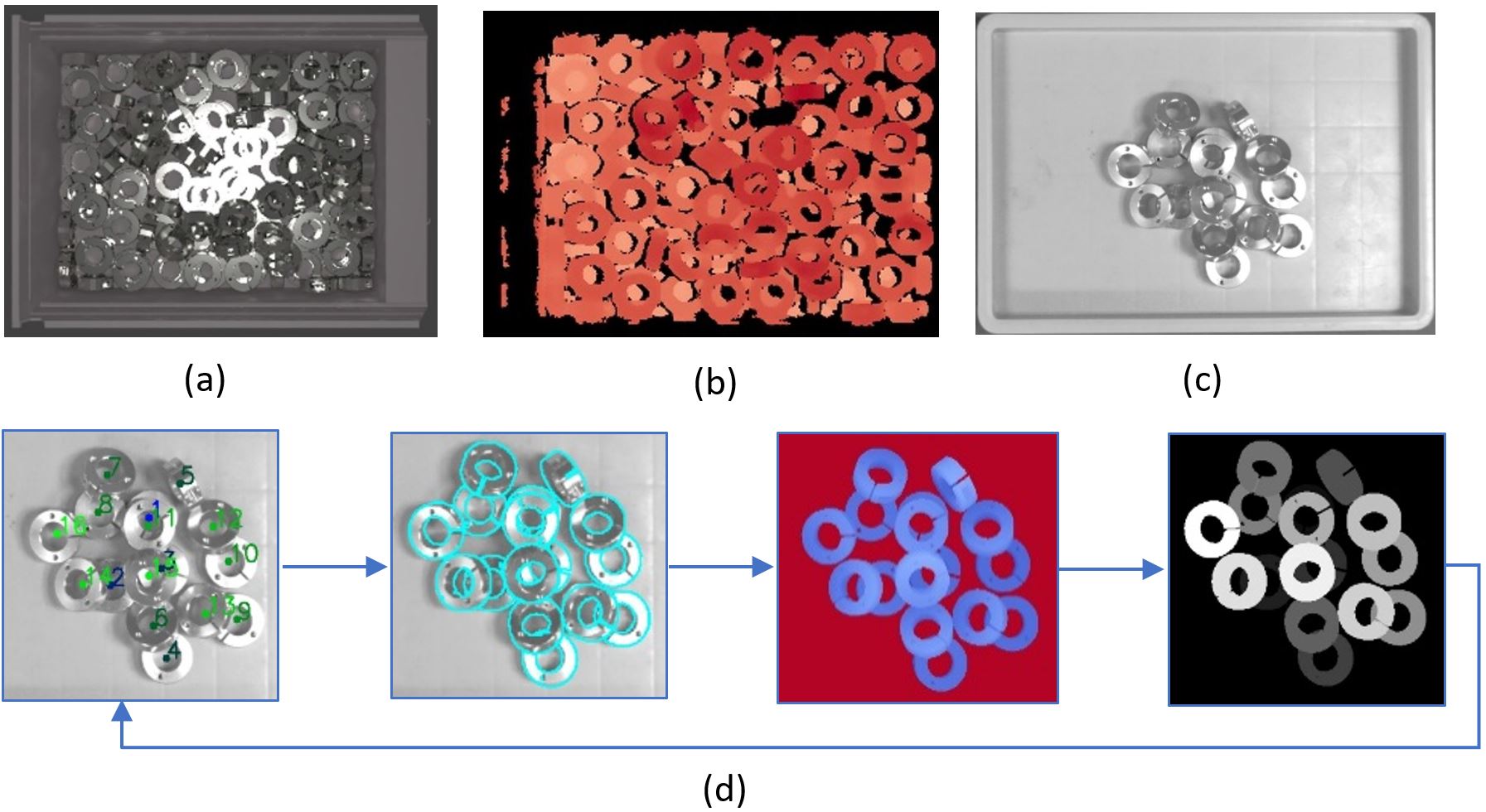}
\end{center}
   \caption{(a) Synthetic image (b) Synthetic depth (c) Real world image (b) Iterative ground truth labeling process on Epson robotics dataset. 
}
\end{figure}

\subsection{Domain Randomization}
To bridge the reality gap and make our model to transfer to the real world, we used domain randomization \cite{DR} by randomly perturbing the following aspects of the scene: (1) texture of background images taken from the Flickr 8K \cite{bg} dataset; (2) position and intensity of point lights, in addition to a planar light for ambient illumination; (3) position and orientation of camera with respect to the scene (pan, tilt, and roll from $-30^\circ$ to $30^\circ$, azimuth from $0^\circ$ to $360^\circ$, elevation from $5^\circ$to $30^\circ$); (4) object material and color with respect to the scene; (5) number and types of depth missing region with respect to input depth map; and (6) number and types of sensor noise with respect to input images. While (1)-(6) are applied to our Epson inhouse dataset, only (5) and (6) are applied on virtual KITTI Dataset.

\section{Experiments}

We conduct extensive experiments on benchmark dataset: virtual KITTI \cite{vkitti} and real KITTI \cite{sparsecnn}. Moreover, to demonstrate the generalization ability, we also perform experiments on our Epson in-house dataset in different scenarios. Note that the best results are marked with \textbf{bold}. $\downarrow$ means smaller is better while $\uparrow$ mean larger is better.

\begin{figure*}[ht]
\begin{center}
\includegraphics[width=0.95\linewidth,height=0.27\textheight]{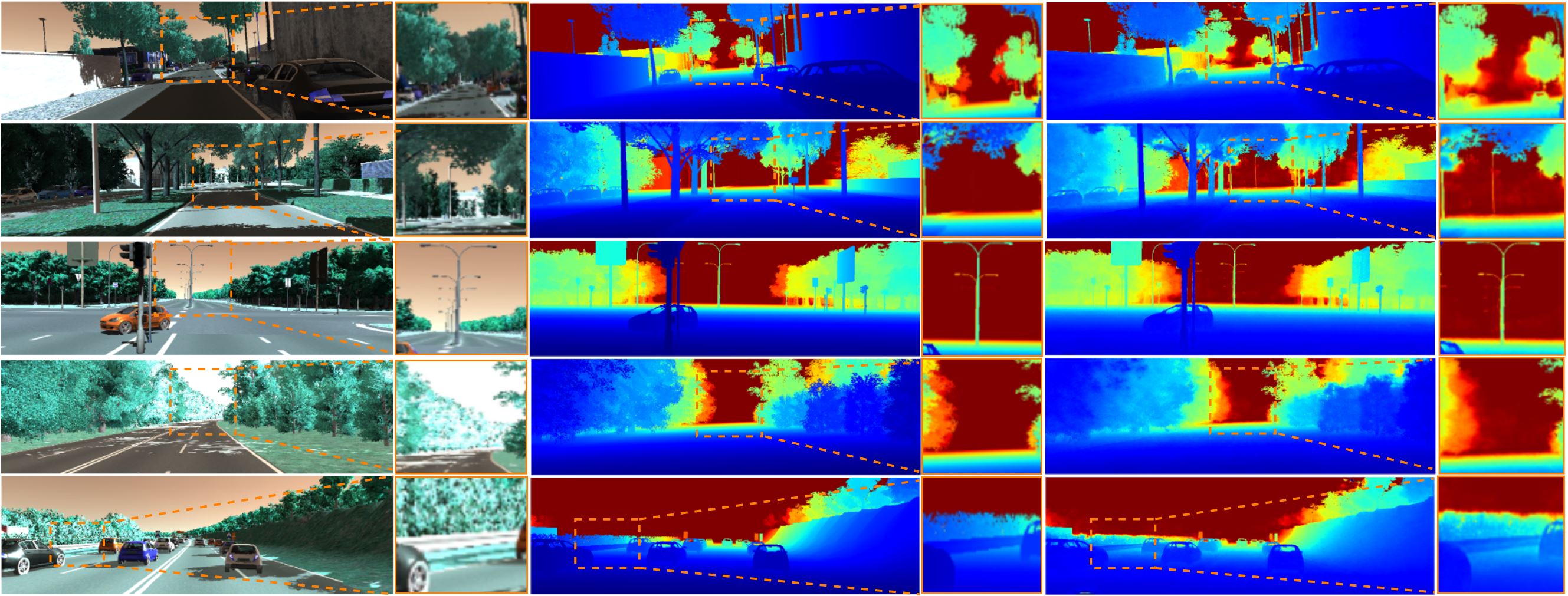}
\end{center}
   \caption{Qualitative results on virtual KITTI datase (5$\%$ sparse depth input). From left to right: image, ground truth depth, and our depth prediction. Compared to ground truth depth, our method is able to predict fine-grained details, learned from fusion features, that even ground truth has not included. }
\end{figure*}

\subsection{Dataset}

\noindent \textbf{Virtual and Real KITTI Dataset} The Virtual KITTI dataset has 21260 stereo pairs in total, including: scene1 (crowded urban area): 4470, scene2 (road in urban area then busy intersection): 2330, scene6 (stationary camera at a busy intersection): 2700, scene18 (long road in the forest with challenging imaging conditions and shadows): 3390, and scene20 (highway driving scene): 8370. In our experiments, the plan to select the train-test split is the following: 1) choose the entire scene6 as test set, 2) choose first 88\% frames from each scene category among the remaining ones for training set, the next 6\% frames for validation set and hence the last 6\% frames for test set, that is resulting in a training set of 16334 images, a validation set of 1113 images, and a testing set of 3813 images. We only use the Real KITTI Datase for evaluation purpose. As there are rare measurement at many areas of images due to LiDAR scans, so the evaluation is set to only count on measurement points.

\noindent \textbf{Epson Robotics Dataset} is composed of synthetic data for training and real world data for testing. We use the PyBullet \cite{pybullet} as our physics simulator and Blender \cite{blender} as our rendering engine to generate synthetic training data as shown in Fig. 4(a) and Fig. 4(b). To be able to evaluate in our real world data, we manually conduct iterative ground truth generation process until the error is less than 1mm, as shown in Fig. 4(d). For each object, the synthetic data contains over 5k training and 1k validation images, and the real test images have about 500 scenes, which is captured in different lighting conditions, bin background colors, bin positions and orientations with a commodity-level RGB-D camera. In our experiment, we have chosen 4 distinct objects that represent shinny, thin and deformable shape scenarios.

\subsection{Evaluation Metrics}
We adopt the standard metrics for the evaluation on virtual KITTI dataset: Root Mean Square Error (RMSE), Mean Absolute Error (MAE). Following the real KITTI benchmark, we also adopt two additional metrics: root mean squared error of the inverse depth (iRMSE) and mean absolute error of the inverse depth (iMAE).

\subsection{Evaluation Performance}
Table \RN{1} shows the quantitative results for depth completion on virtual KITTI dataset. Because the sparsity of depth points increase with the range, the results present the performance in the range of 20 m, 50 m and 100 m, respectively. Table \RN{1} shows our method outperforms SOTA in both range of 0-50m and 0-100m, while it is comparable to SOTA in 0-20m range. As shown by Fig. 5, our method enhance depth estimation with fine-grained details, that even ground truth has not included. Therefore, the predicted depth value at fine-grained region is expected to be different from ground truth value. Consequently, the average error between ground truth and predictions has been increased in these regions.   

\begin{table}[ht]
\centering
\begin{adjustbox}{width=0.45\textwidth}
\small
\begin{tabular}{c|cc|cc|cc}
  \hline
  & \multicolumn{2}{c|}{0-20 (cm)} &\multicolumn{2}{c|}{0-50 (cm)}  &  \multicolumn{2}{c}{0-100 (cm)}  \\ 
  Method& MAE $\downarrow$ & RMSE$\downarrow$ & MAE $\downarrow$ & RMSE$\downarrow$  & MAE $\downarrow$ & RMSE$\downarrow$  \\ 
   \hhline{=|==|==|==}
  MRF \cite{mrf} & 56.67 & 116.776 & 131.03 & 312.41 & 209.45 & 575.20 \\ 
  TGV \cite{tgv} & 41.85 & 114.57 & 113.38 & 323.97& 205.78& 621.48  \\ 
  STD\cite{sparsedense} & 258.98 & 386.91 & 653.54 & 1066.55 &  1072.52 & 1892.04\\ 
  SCNN \cite{sparsecnn} & 56.44 &	137.34 & 153.01 & 384.96 & 258.23 & 681.13 \\ 
  \hline
  Ours & 22.46 & \textbf{84.77} &  \textbf{61.03} &  \textbf{211.68} &   \textbf{78.19} &  \textbf{271.21}  \\ 
   \hline
\end{tabular}
\end{adjustbox}
\caption{Quantitative comparison with methods on virtual KITTI dataset.} 
\end{table}

\begin{figure*}[ht]
\begin{center}
\includegraphics[width=0.95\linewidth,height=0.27\textheight]{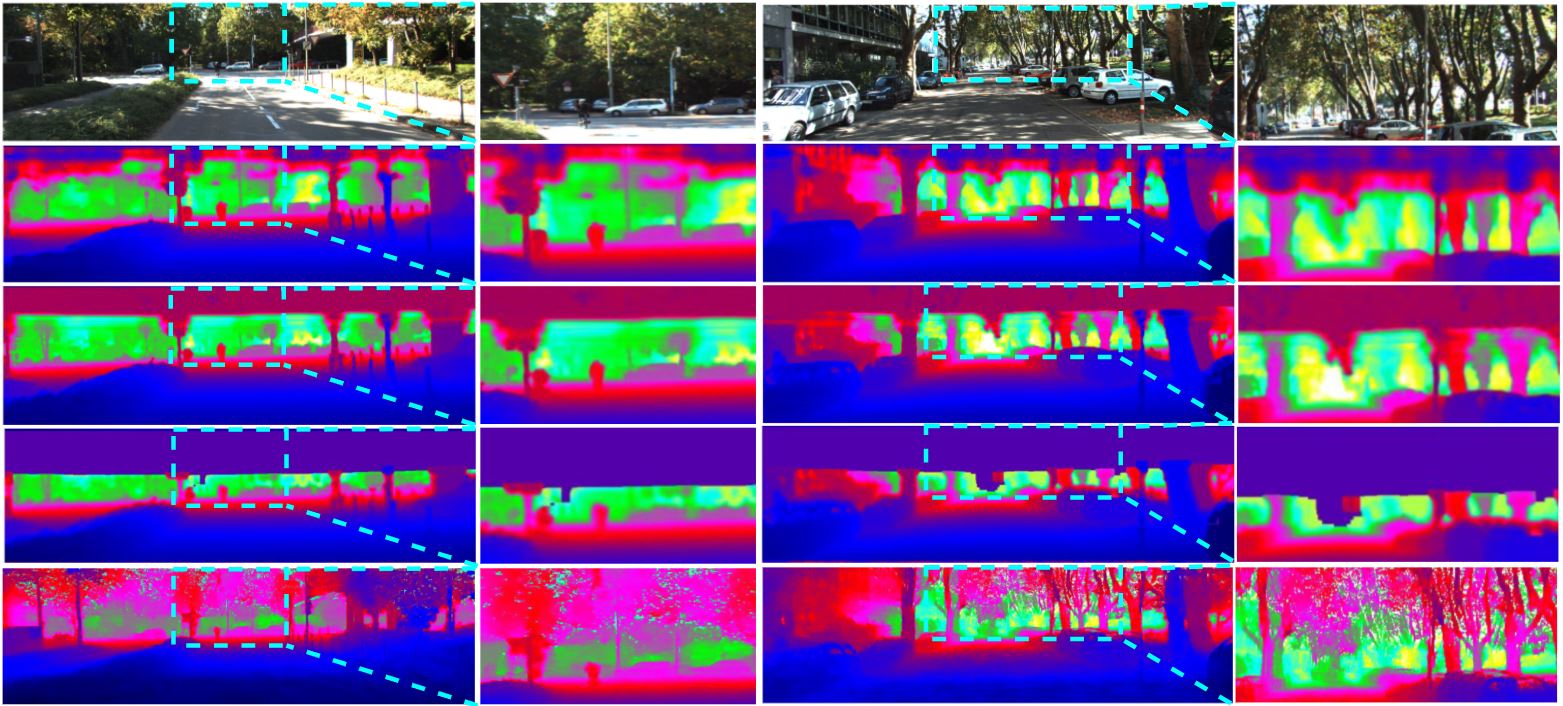}
\end{center}
   \caption{Qualitative results on real KITTI dataset. From top to bottom: DFuseNet \cite{dfuse} Morph-Net \cite{morph}, SparseConv \cite{sparsecnn} and Ours. As we can see from the zooming region, our method is able to recover depth values for fine-grained details even at the far end.}
\end{figure*}

To demonstrate the generalization of our proposed method quantitatively in real world scenario, we also test our method on the real KITTI dataset as presented in Table \RN{2}. Different from other methods which train on real KITTI dataset, we train on virtual KITTI dataset only. As can be seen from Table \RN{2}, our proposed method has achieved very close to SOTA, with 4.01\% below for RMSE, 4.91\% below for iRMSE, and 2.5\% below for iMAE. Nevertheless, we outperform SOTA for MAE by 9.81\%. This is because we enforce our model for instance aware consistency by joint tasks learning in order to regularize the depth estimation and mitigate the impact of noise. Our model also recover fine-grained details for depth completion as shown in Fig. 6. Because the ground truth of real KITTI dataset also has not demonstrated this fine-grained details due to sparse measurement of LiDAR scans, higher error between ground truth and predictions in these areas are also expected.

\begin{table}[ht]
\centering
\begin{adjustbox}{width=0.45\textwidth}
\small
\begin{tabular}{c|c|c|c|c}
  \hline
  Method& RMSE (cm) $\downarrow$ & MAE (cm) $\downarrow$ & iRMSE (1/km) $\downarrow$ & iMAE (1/km) $\downarrow$  \\ 
   \hhline{=|=|=|=|=}
  DFuse \cite{dfuse}  & 120.66 & 42.99 & 3.62 & 1.79  \\ 
  MorNet \cite{morph}   & 104.54 & 31.04 & 3.84 & 1.57 \\ 
  CSPN \cite{cspn}  & 101,96  & 27.94 & 2.93  & 1.15  \\ 
  HMSNet \cite{hms}  & 93.75 & 25.84 & 2.93 & 1.14  \\ 
  NConv \cite{nconv} & 82.99 & 23.32 & 2.60 & \textbf{1.03}  \\ 
  STD \cite{cnn14}  & 81.47 & 24.99 & 2.80  & 1.21  \\ 
  DNorm \cite{depthnormal} & \textbf{77.71} & 23.51 & \textbf{2.42} & 1.13  \\ 
  \hline
  \makecell{Ours} & 81.01 & \textbf{21.41} & 2.55 & 1.16   \\ 
   \hline
\end{tabular}
\end{adjustbox}
\caption{Quantitative comparison with methods on real KITTI dataset. Note that our method train with synthetic data only} 
\end{table} 

To show the benefits for our robotics manipulation tasks, we also evaluate the performance on our Epson in-house robotics dataset as shown in Fig. 7. From the Table \RN{3}, our network improve the sensor depth quality significantly. Overall, we improve MAE by 24.8\% and RMSE by 16.8\% in non-missing region, while improve 1.96\% for MAE and 9.95\% for RMSE in missing region.

\begin{table}[ht]
\centering
\begin{adjustbox}{width=0.4\textwidth}
\begin{tabular}{c|cc|cc|cc}
  \hline
   \multirow{3}{*}{Object}   & \multicolumn{4}{c|}{Non-missing Region} & \multicolumn{2}{c}{Missing Region} \\
   \cline{2-7}
  & \multicolumn{2}{c|}{Sensor (mm)} & \multicolumn{2}{c|}{Ours(mm)} & \multicolumn{2}{c}{Ours(mm)}  \\

  & MAE & RMSE & MAE & RMSE  & MAE & RMSE \\ 
  \hhline{=|==|==|==}
  Object 1 & 2.74 & 4.39 & 2.15 & 3.42  & 2.36& 3.10  \\ 
  Object 2 & 1.43 & 2.54 & 0.77 & 2.17 & 1.54 & 2.60 \\ 
  Object 3 & 2.44 & 4.13 & 1.96 & 3.79& 2.56 & 4.42  \\ 
  Object 4 & 0.53 & 1.11 & 0.49 & 0.74 & 0.54 & 0.83 \\
  \hline
\end{tabular}
\end{adjustbox}
\caption{Performance evaluation on Epson robotics dataset} 
\end{table}

\subsection{Ablation Study}
For better understanding of our approach, we explore ablation studies on how each module boost the performance for depth completion task. We remove all modules and denote the depth prediction only as our baseline. The experimental results Table \RN{4} show that after integrating with all modules, the performance of depth completion present the best promising results. This highlights a
major advantage of instance aware learning, which strengthen the model's
ability to predict depth values based on pixel-wise semantic label of instance features.

\begin{table}[ht]
\centering
\begin{adjustbox}{width=0.48\textwidth}
\small
\begin{tabular}{l|c|c|c|c}
  \hline
  Method& RMSE (cm) $\downarrow$ & MAE (cm) $\downarrow$ & iRMSE (1/km) $\downarrow$ & iMAE (1/km) $\downarrow$  \\ 
   \hhline{=|=|=|=|=}
  Baseline  & 224.31 & 33.89 & 3.68 & 1.86  \\ 
  + stack   & 161.81 & 31.02 & 3.54 & 1.77 \\ 
  + aspp  & 113.96  & 27.44 & 2.91  & 1.65  \\ 
  + self-guided  & 97.21  & 26.16 & 2.73  & 1.36  \\
  \hline
  Full w/o IA-L & 92.31 & 24.80 & 2.65 & 1.33  \\
  Full w/ IA-L & \textbf{81.01} & \textbf{21.41} & \textbf{2.55} & \textbf{1.16}  \\
  \hline
\end{tabular}
\end{adjustbox}
\caption{Ablation study: performance changes on real KITTI dataset with each component of our proposed model.} 
\end{table}

\begin{figure}[ht]
\begin{center}
\includegraphics[width=0.99\linewidth,height=0.3\textheight]{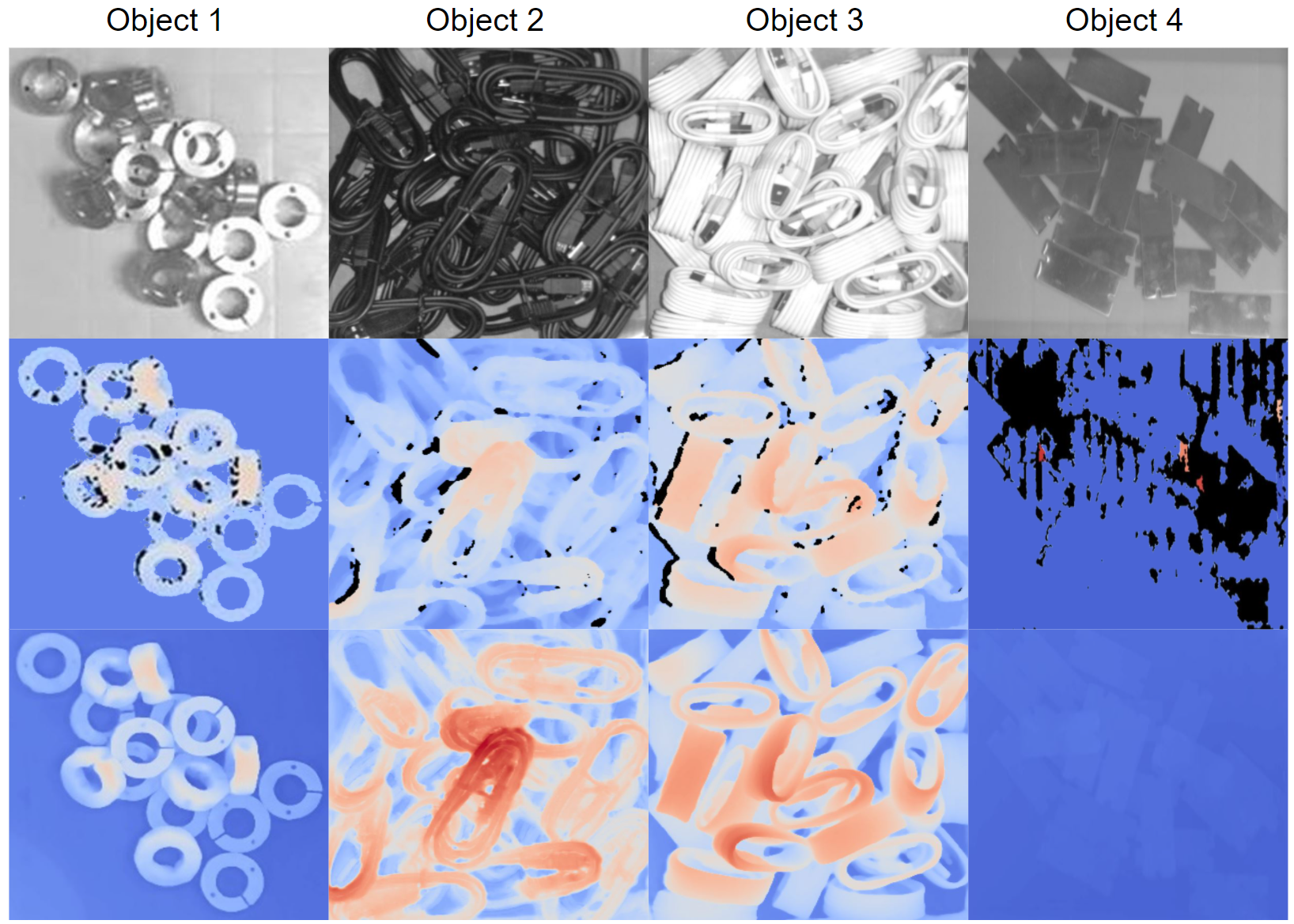}
\end{center}
   \caption{Qualitative results on Epson robotics dataset. From top to bottom: grayscale image, original low resolution depth, and ours. From our results, our model not only predict missing depth, but also recover fine-grain details.
}
\end{figure}

\section{Conclusion}
In this work, we present a novel end-to-end self-guided instance-aware network (SG-IANet) for depth completion and enhancement. We introduce geometry and context learning, especially for occluded or truncated objects, to predict clearer and sharper structures of object. In addition, we also introduce instance-aware learning and show mutual improvements. Incorporate with muti-scale levels of feature maps from image branch of model, we show that the self-guided module allows the network to learn to focus on meaningful features for accurate depth completion. Last but not least, we propose to utilize domain randomization to improve robustness and generalization to unseen environment. Extensive experiments demonstrate that our proposed method outperforms on synthetic dataset, and comparable performance on real dataset though our model is trained on synthetic only. Moreover, ablation studies validate the effectiveness of proposed modules.

\bibliographystyle{IEEEtran}
\bibliography{ref}

\end{document}